\documentclass[times, review, 10pt]{elsarticle}
\usepackage{lineno,hyperref}
\usepackage{array}
\usepackage{amssymb}
\usepackage{enumerate}
\usepackage{makecell}
\usepackage{mathtools}
\usepackage{array}
\usepackage{multirow}
\usepackage{ulem}
\usepackage{color}
\usepackage{booktabs}
\usepackage{diagbox}
\usepackage{enumerate}
\usepackage{graphicx}
\usepackage{subfigure}
 \usepackage{tikz,mathpazo} 
\usepackage{indentfirst}
\usepackage{amssymb}  
\usepackage{pifont}   
\newcommand{\xmark}{\ding{55}}  
\usepackage{soul}
\usepackage{amsmath}
\usepackage[justification=centering]{caption}
\usepackage{tabularx}
\usepackage{algorithm}
\usepackage{algorithmicx}
\usepackage{algpseudocode}
\usepackage{adjustbox}
\usepackage{geometry}
\usetikzlibrary{shapes.geometric, arrows}
\modulolinenumbers[5]

\soulregister{\cite}7
\soulregister{\ref}7

\newcommand{\circledsmall}[1]{\lower.7ex\hbox{\tikz\draw (0pt, 0pt)%
    circle (.5em) node {\makebox[0.1em][c]{\small#1}};}}

\newcommand{\circledtiny}[1]{\lower.7ex\hbox{\tikz\draw (0pt, 0pt)%
    circle (.3em) node {\makebox[0.1em][c]{\tiny #1}};}}

\journal{Journal of \LaTeX\ Templates}

\begin{document}

\begin{frontmatter}

\title{Complementarity-driven Representation Learning for Multi-modal Knowledge Graph Completion}

    \author[label5]{Lijian Li}

    \affiliation[label5]{organization={Department of Computer and Information Science, University of Macau},
    	city={Macau},
    	postcode={999078},
    	country={China}}

\begin{abstract}
Multi-modal Knowledge Graph Completion (MMKGC) aims to uncover hidden world knowledge in multimodal knowledge graphs by leveraging both multimodal and structural entity information. However, the inherent imbalance in multimodal knowledge graphs, where modality distributions vary across entities, poses challenges in utilizing additional modality data for robust entity representation. Existing MMKGC methods typically rely on attention or gate-based fusion mechanisms but overlook complementarity contained in multi-modal data. In this paper, we propose a novel framework named Mixture of Complementary Modality Experts (MoCME), which consists of a Complementarity-guided Modality Knowledge Fusion (CMKF) module and an Entropy-guided Negative Sampling (EGNS) mechanism. The CMKF module exploits both intra-modal and inter-modal complementarity to fuse multi-view and multi-modal embeddings, enhancing representations of entities. Additionally, we introduce an Entropy-guided Negative Sampling mechanism to dynamically prioritize informative and uncertain negative samples to enhance training effectiveness and model robustness. Extensive experiments on five benchmark datasets demonstrate that our MoCME achieves state-of-the-art performance, surpassing existing approaches.
\end{abstract}


\end{frontmatter}

\section{Introduction}
Knowledge graphs (KGs) \cite{bordes2013translating, chao2020pairre, sun2019rotate, trouillon2016complex, hu2024knowledge} model real-world knowledge through structured representations in the form of triples—comprising a head entity, a relation, and a tail entity—which are typically constructed manually based on existing databases. However, the inherent incompleteness of KGs \cite{wang2022entity, zheng2021knowledge}, coupled with the high cost of annotating factual triples, has given rise to the task of Knowledge Graph Completion (KGC), which aims to predict and infer missing but plausible triples within an existing knowledge graph. Conventional KGC methods \cite{bordes2013translating, chao2020pairre, sun2019rotate, trouillon2016complex} predominantly rely on Knowledge Graph Embedding (KGE) techniques, where entities and relations are embedded into continuous vector spaces to learn structural representations that model the relational patterns of triples and assess their plausibility. 

Conventional KGC methods can generally be categorized into translation-based and semantic matching approaches. Translation-based methods (e.g., \cite{bordes2013translating, sun2019rotate, chao2020pairre}) treat triple modeling as a translation operation from the head to the tail entity, using distance-based scoring functions to evaluate the plausibility of a triple. In contrast, semantic matching methods (e.g., \cite{trouillon2016complex}) focus more on the intrinsic compatibility between entities and relations, often employing tensor factorization frameworks to design matching mechanisms that capture latent structural dependencies. Additionally, recent studies have explored the utilization of neural architectures, including convolutional neural networks (CNNs) and graph neural networks (GNNs) (e.g., \cite{DBLP:conf/aaai/ShangTHBHZ19, DBLP:conf/naacl/NguyenNNP18, DBLP:conf/iclr/VashishthSNT20}), to model the complex interactions between entities and relations. These models aim to automatically extract structural and semantic features, thereby enhancing the expressiveness of learned representations. Nonetheless, traditional KGC models are primarily limited to exploiting relational information in the KG and often neglect the rich multimodal information, such as textual, visual, and numerical attributes—embedded within entities.

Multimodal Knowledge Graphs (MMKGs) \cite{cai2017kbgan, zhang2023modality, xu2022relation, zhang2022knowledge, cao2022otkge} have emerged as a significant extension of conventional KGs by incorporating diverse modality-specific attributes (e.g., videos, audios, and numerical data), which enrich the semantic representations of entities. MMKGs have become a foundational component in numerous AI applications, particularly in recommender systems \cite{wang2019kgat} and natural language processing. Correspondingly, the task of Multimodal Knowledge Graph Completion (MMKGC) has also gained increasing attention.

Existing MMKGC methods \cite{cai2017kbgan, zhang2023modality, li2023imf, lee2023vista, zhang2022knowledge} typically incorporate multimodal information as auxiliary modality embeddings, which are integrated into the entity representation space to enhance the expressiveness of the learned embeddings. However, several challenges persist in real-world scenarios. A key issue lies in the uneven distribution of modalities across entities, whereby certain entities may lack essential modalities due to data sparsity. To address this, the NATIVE framework \cite{zhang2024native} introduces a relation-guided weighting mechanism for optimizing multimodal fusion, along with an adversarial training strategy to mitigate the imbalance across modalities. Nevertheless, conventional fusion strategies—such as simple concatenation or attention-based mechanisms (e.g., \cite{DBLP:conf/ijcai/XieLLS17, DBLP:conf/mm/ZhangQFX19})—often fall short in capturing the fine-grained intra-modality dynamics and context-dependent inter-modality complementarity among multimodal features. This limitation is especially pronounced in practical applications, where some modalities may be missing, incomplete, or noisy. In such cases, explicitly modeling the collaborative and compensatory mechanisms between modalities becomes crucial. Recent work \cite{limodality} has demonstrated that when one or more modalities are unavailable or corrupted, effectively leveraging modality complementarity can significantly mitigate performance degradation. The finding further substantiates the pivotal role of modality complementarity in enhancing the robustness and fusion efficacy of multimodal learning systems. Additionally, most negative sampling-based KGE methods tend to assume uniform importance across all negative samples during training. However, the assumption fails to account for the intrinsic heterogeneity of negative samples in multimodal settings, where semantic richness, modality characteristics, and similarity to positive samples may vary widely. Treating all negatives equally may lead the model to overfit on trivial or semantically irrelevant samples, thereby impairing its ability to distinguish critical semantic relationships and ultimately reducing the discriminative power and generalization performance of the learned embeddings.

To address the limitations of existing multimodal knowledge graph completion (MMKGC) methods including insufficient modeling of modality complementarity, imbalanced distribution of modality information across entities, and overly simplistic negative sampling strategies, we propose a novel unified framework for Mixture of Complementary Modality Experts \allowbreak{} (MoCME). Built upon the Mixture of Experts (MoE) paradigm, our MoCME is designed to comprehensively exploit the synergy and complementarity across modalities, thereby enabling more expressive and robust entity representations. The framework consists of two key components: 1) Complementarity-
guided Modality Knowledge Fusion (CMKF): it contains a Modality-specific Expert module and a Cross-modality Fusion module. For each modality, we employ a set of parallel expert networks to process its pre-extracted embeddings. Each expert is specialized to capture different semantic subspaces, and their outputs are regarded as multi-view representations of the entity within that modality. To fully exploit intra-modality complementarity, an adaptive view-level aggregation mechanism is introduced, which dynamically integrates diverse expert outputs based on their internal correlations, resulting in rich and fine-grained modality-specific representations. To synthesize information across different modalities, we introduce a modality-level fusion strategy guided by inter-modality complementarity. The mechanism adaptively integrates multiple modality embeddings to form a unified multimodal representation of the entity. The design enhances the model’s robustness and generalization, particularly in scenarios where some modalities are missing, incomplete, or corrupted. 2) Entropy-guided Negative Sampling (EGNS): to model relational semantics, we adopt the RotatE scoring function, which effectively captures entity-relation interactions through complex space rotations. Furthermore, we propose an entropy-based negative sampling strategy to address the limitations of uniform negative sampling, which dynamically assigns greater importance to more uncertain and informative negative samples, encouraging the model to focus on semantically challenging cases and improving its discriminative capacity.

\begin{itemize}
    \item We propose a novel \textbf{MoCME} framework that jointly models multi-view semantics within each modality and inter-modality synergy across modalities, achieving comprehensive and robust multimodal representations through expert-based adaptive fusion.
    
    \item We introduce a Complementarity-guided Modality Knowledge Fusion (CMKF) module that operates at both intra- and inter-modality levels, enabling fine-grained information integration and effective handling of incomplete or noisy modalities.
    
    \item We develop an Entropy-guided Negative Sampling (EGNS) strategy, which prioritizes harder negatives and improves the model’s ability to learn from informative samples, thereby enhancing robustness and generalization.
    
    \item Extensive experiments on widely-used benchmarks demonstrate that our MoCME achieves state-of-the-art performance, validating its effectiveness in multimodal knowledge representation and completion tasks.
\end{itemize}

\section{Related works}
\subsection{Knowledge Graph Completion}
Knowledge graph completion \cite{liang2024survey} is a fundamental task that involves discovering missing triples in a given knowledge graph. Traditional KGC approaches, also known as knowledge graph embedding  \cite{wang2017knowledge} methods, map entities and relations to continuous vector spaces. These models utilize distinct score functions to evaluate the likelihood of triples, aiming to assign higher scores to valid triples and lower scores to invalid ones. KGE models primarily fall into two categories: translation-based methods, which view relations as translations between entities (e.g., TransE \cite{bordes2013translating}, RotatE \cite{sun2019rotate}, PairRE \cite{chao2020pairre}), and semantic matching methods, which leverage similarity measures through tensor decomposition (e.g., DistMult \cite{yang2014embedding}, ComplEx \cite{trouillon2016complex}, TuckER \cite{balavzevic2019tucker}). To be specific, TransE \cite{bordes2013translating} models relations as translations in the embedding space by assuming that the tail entity embedding should be close to the head entity embedding plus the relation vector, i.e., $\mathbf{h} + \mathbf{r} \approx \mathbf{t}$. RotatE \cite{sun2019rotate} models relations as rotations in the complex vector space, where each relation is represented as a phase rotation, and the head and tail entities are embedded as complex vectors such that $\mathbf{t} \approx \mathbf{h} \circ \mathbf{r}$, where $\circ$ denotes the Hadamard (element-wise) product. DistMult \cite{yang2014embedding} models the plausibility of a triple by computing a bilinear dot product between the head, relation, and tail embeddings, but it is inherently limited to modeling symmetric relations due to the commutativity of multiplication. ComplEx \cite{trouillon2016complex} extends DistMult into the complex vector space, enabling the representation of asymmetric relations by leveraging complex-valued embeddings and Hermitian dot products.

Multi-modal knowledge graph completion builds upon traditional KGC by incorporating multi-modal information from multi-modal knowledge graphs, such as visual, textual, numeric and other modalities, which represent diverse perspectives of entities. MMKGC methods can be classified into three categories: Modal Fusion Methods \cite{wang2021visual, xie2016image}, Modal Ensemble Methods \cite{li2023imf, cao2022otkge} and Negative Sampling (NS) Enhanced Methods \cite{cai2017kbgan, zhang2023modality, xu2022relation, zhang2022knowledge}. Specifically, Xie et al. \cite{xie2016image} propose a visual-enhanced knowledge representation model that integrates image features into entity embeddings via joint learning, enabling the model to capture complementary visual semantics alongside structured knowledge. Li et al. \cite{li2023imf} introduce IMF, a modality-aware ensemble framework that leverages both modality-specific experts and a global aggregator to dynamically balance and combine different modal representations. Cai and Wang \cite{cai2017kbgan} design KBGAN, an adversarial training approach that selects informative negative samples by using a generative model to guide the discriminator, improving training efficiency and embedding quality. Zhao et al. \cite{xu2022relation} propose a relation-enhanced negative sampling strategy that incorporates relational context into the selection of negative triples, enabling the model to distinguish harder negatives and improving the performance of multimodal knowledge graph completion. Despite the above progress, existing MMKGC methods often overlook the role of modality complementarity in multimodal feature fusion. In addition, issues such as data imbalance and the complexity of multimodal data distributions remain underexplored. Our work aims to explicitly explore the role of complementarity in MMKGC.

\subsection{Complementarity-based Multi-modal Learning}
Complementarity in multimodal data is crucial for addressing modality imbalance by adaptively weighting and fusing features, allowing the model to rely on richer modalities to compensate for missing or noisy information in others. Recent studies have increasingly emphasized the value of complementarity-driven fusion in enhancing robustness and generalization in multimodal tasks \cite{zhang2024multimodal, wanyan2023active, limodality}. Wanyan et al. \cite{wanyan2023active} propose an active multimodal few-shot action recognition framework that explicitly models the task-dependent complementarity between modalities. Similarly, Li et al. \cite{limodality} empirically demonstrate that complementarity is critical for performance under missing or noisy modalities and advocate treating it as a controllable variable.

To improve multimodal fusion under uncertainty, a variety of methods based on generalized evidence theory, probabilistic modeling, and belief functions have been introduced. He et al. explore the fundamental issues of uncertainty and evidence conflict by designing new base functions and combination rules in belief theory \cite{he2022new, he2021conflicting}. Extensions such as MMGET \cite{he2022mmget} and TDQMF \cite{he2023tdqmf} integrate Markov models and quantum theory into evidential reasoning, enabling the fusion process to better handle ambiguous or partially missing modality inputs. To quantify and regulate uncertainty in fused representations, novel entropy formulations such as Ordinal Belief Entropy \cite{he2023ordinal} and Ordinal Fuzzy Entropy \cite{he2022ordinal} have been proposed. These methods are particularly useful in scenarios requiring conflict-sensitive decision-making and have shown promise in multi-modal settings.

In addition to theoretical modeling, these evidential principles have been widely applied to real-world domains such as medical image analysis and knowledge graph learning. For example, uncertainty-aware models for semi-supervised medical image segmentation—including EPL \cite{he2024epl}, Mutual Evidential Deep Learning \cite{he2024mutual}, Uncertainty-Aware Fusion \cite{he2024uncertainty}, Co-Evidential Fusion \cite{he2025co}, and Prototype Consistency Learning \cite{li2024efficient}—demonstrate the benefits of explicitly modeling evidential support and modality complementarity. Additionally, matrix-based distance functions for fuzzy set reasoning \cite{he2021matrix} and multi-view clustering frameworks leveraging conditional entropy \cite{li2025adaptive} have also been utilized to enhance complementarity-driven learning in structured and unstructured modalities.

Complementarity-aware designs have also been extended to knowledge graph completion. Recent frameworks such as Unitrans \cite{huang2025unitrans} promote knowledge transfer across vertical domains by balancing multi-source evidence, while Multi-Prototype Embedding Refinement \cite{bi2025multi} and Multi-View Riemannian Fusion \cite{li2025multi} further improve entity-level representation learning through coordinated multimodal information. Complementarity has also been recognized in aspect-based sentiment analysis \cite{li2022nndf} and temporal action localization \cite{he2024generalized}, where weak supervision and noisy inputs demand evidence-aware learning mechanisms. Even in spatiotemporal remote sensing tasks \cite{xu2023spatio}, complementarity between image features and vegetation indices has been used to improve classification accuracy.

In foundational KG completion tasks, regularization approaches \cite{li2025rethinking} and structure-aware models \cite{li2025towards} have also started to exploit modality interactions and uncertainty modeling. The benefit of belief-based reasoning extends to code understanding as well, as evidenced by self-debugging frameworks \cite{chen2025revisit} incorporating uncertainty-aware feedback mechanisms.

Motivated by these diverse yet consistent findings \cite{wanyan2023active, limodality, he2022new, he2023ordinal, he2022mmget, he2024generalized, he2025co}, our work integrates complementarity modeling into the multimodal knowledge graph completion pipeline. By explicitly modeling the conflict, reliability, and informativeness of different modalities, we aim to build a fusion framework that not only balances contributions adaptively but also leverages complementary signals to enhance reasoning under uncertainty.

\section{Methodology}
\begin{figure*}
    \centering
    \includegraphics[scale=1.0]{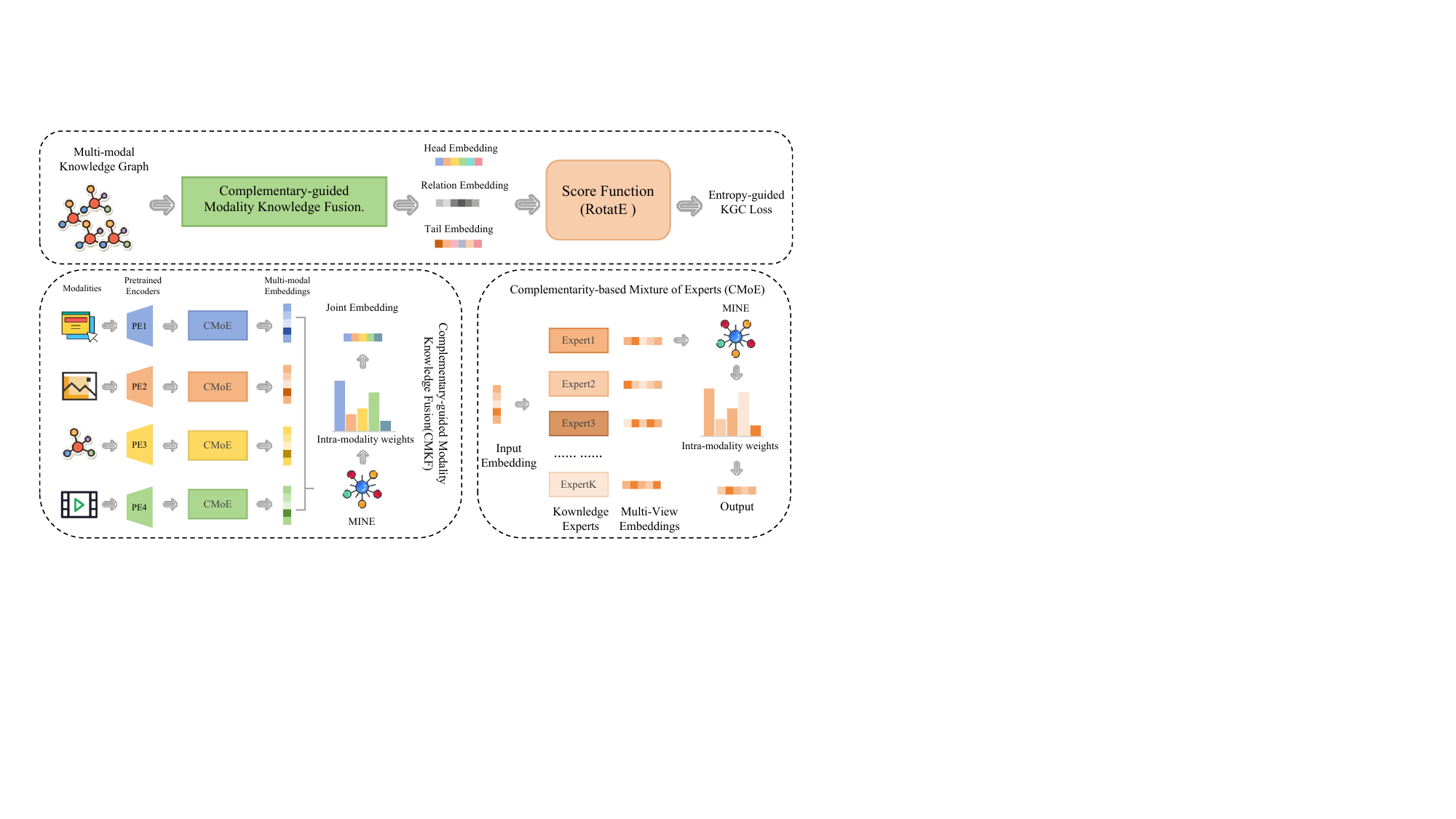}
    \caption{The overview of the proposed MoCME framework, which consists of two main modules: a Complementary-guided Modality Knowledge Fusion module and an entropy-based KGC loss. The complementary-guided modality knowledge fusion module leverages intra-modal and inter-modal complementarity as adaptive weights to generate a unified multimodal embedding for knowledge graph completion. A complementarity-based mixture of experts (CMoE) module produces multi-view embeddings for each modality, which are fused using intra-modal complementarity to enhance representation. The resulting modality embeddings are then integrated using inter-modal complementarity for a comprehensive representation. Moreover, an entropy-weighted contrastive loss is designed to focus on harder negative samples generated by the negative sampling strategy, thereby improving the performance and robustness of the model.}
    \label{fig:enter-label}
\end{figure*}

In this section, we present a comprehensive overview of our proposed framework, Mixture of Complementary Modality Experts (MoCME), which is designed to address the challenges of multimodal knowledge graph completion through two key components: (1) a Complementarity-guided Modality Knowledge Fusion (CMKF) module including a Modality-specific Expert Module and a Cross-modality Fusion module, which integrates multi-view and multi-modal embeddings by leveraging both intra- and inter-modality complementarity to construct robust and expressive entity representations; and (2) an Entropy-guided Negative Sampling (EGNS) mechanism, which dynamically prioritizes informative and uncertain negative samples to enhance training effectiveness and model robustness. Furthermore, to assess the plausibility of knowledge triples, we employ the RotatE scoring function, which enables the modeling of various complex relational patterns through phase rotations in the complex vector space. Fig. \ref{fig:enter-label} illustrates the overall architecture and workflow of the proposed MoCME framework. Next, we will elaborate on the MoCME framework in terms of its architecture design and technical details.

\subsection{Task Definition}
We define a multi-modal knowledge graph (MMKG) as $\mathcal{MG} = (\mathcal{E}, \mathcal{R}, \mathcal{T}, \mathcal{M})$, where $\mathcal{E}$ and $\mathcal{R}$ denote the sets of entities and relations, respectively. The triple set $\mathcal{T} = {(h, r, t) \mid h, t \in \mathcal{E}, r \in \mathcal{R}}$ encodes the structural knowledge in the form of relational facts, indicating that a head entity $h$ is connected to a tail entity $t$ via relation $r$. In addition to structural triples, MMKGs are enriched with a modality set $\mathcal{M}$, which consists of various types of auxiliary information such as images, textual descriptions, numerical attributes, audio signals, or videos, each capturing complementary semantic aspects of the entities. For a given entity $e \in \mathcal{E}$ and a specific modality $m \in \mathcal{M}$, we denote its modality-specific information as $\mathcal{I}_m(e)$, which may differ in format and feature space across modalities. Importantly, the structural component $\mathcal{T}$ can also be viewed as a distinct modality, capturing relational semantics that are often complementary to non-relational data. The unified formulation allows multimodal reasoning frameworks to jointly model the interactions across structural and non-structural modalities. To assess the plausibility of a given triple $(h, r, t)$, MMKGC models typically employ a score function $\mathcal{S}(h, r, t)$, where higher scores reflect a greater likelihood that the triple expresses a valid fact. In our work, we adopt RotatE as the structural scoring function, which models relations as rotations in the complex vector space, which allows it to effectively capture diverse relational patterns such as symmetry, antisymmetry, inversion, and composition, making it well-suited for structural reasoning in MMKGs.

\subsection{Complementary-guided Modality Knowledge Fusion}
The critical role of modality complementarity in improving the robustness and generalization of multimodal learning has been well established in prior studies \cite{limodality}. Complementary modalities often carry non-overlapping yet semantically relevant information, enabling models to better cope with missing, noisy, or unreliable inputs. Motivated by this insight, we propose a Complementarity-guided Modality Knowledge Fusion (CMKF) framework that explicitly models both intra-modality and inter-modality complementarity to enhance entity representation learning in MMKGs.

To begin with, we perform modality encoding to extract semantic features from various modalities associated with each entity. This step lays the foundation for multimodal representation learning in MMKGC and is widely adopted in prior works. For widely-used visual and textual modalities, we employ pre-trained encoders $PE_m$ to obtain the initial raw feature vector for modality $m$.

Given a modality $m$ and an entity $e \in \mathcal{E}$, we denote the set of raw modality data associated with $e$ as $\mathcal{I}_m(e) = \{\mathcal{I}_{e,m}^{(1)}, \mathcal{I}_{e,m}^{(2)}, \dots\}$, where each $\mathcal{I}_{e,m}^{(i)}$ represents an individual modality-specific element (e.g., an image, a sentence, or a numeric token). The raw modality representation $\textbf{v}_m^r$ is then computed by averaging the encoder outputs over all modality elements:

\begin{equation}
\textbf{v}_m^r = PE_m(\mathcal{I}_m(e))
\end{equation}

where the encoders $PE_m$ are pre-trained on large-scale external datasets and remain frozen during training to preserve their generalization capabilities and reduce training complexity. Specifically, we utilize VGG16 \cite{simonyan2014very} for visual information and BERT \cite{devlin2018bert} for textual inputs, which ensures that each modality retains its high-level semantic characteristics and avoids overfitting within the limited MMKG data.

Moreover, in order to bring different modalities into a unified representation space, the raw modality feature $\textbf{v}_m^r$ is further passed through a projection network $f_m$, which consists of a two-layer multilayer perceptron (MLP) with ReLU activation. The resulting modality-specific embedding is denoted as $\mathbf{v}_m = f_m(\textbf{v}_m^r) \in \mathbb{R}^{d_e}$, where $d_e$ is the unified embedding dimension across all modalities.

In addition to typical visual and textual modalities, we extend the structural modality—originally encoded as symbolic triples—into token sequences and encode them using BERT, treating structure as another semantic-rich modality. Likewise, for numeric modality, we convert scalar values into sequences via tokenization and encode them with BERT as well, enabling them to benefit from the same semantic modeling framework.

However, unlike other modalities, the final structural embedding $\mathbf{e}_S$ is not directly extracted from a frozen encoder but rather initialized as a learnable parameter. It is optimized jointly with the KG completion objective to better fit the triple-based relational structure, and thus serves as the backbone modality in the overall representation.

To capture fine-grained intra-modality variation and enhance expressiveness, we adopt a mixture-of-experts design inspired by MoMoK \cite{zhang2024mixture}. For each modality $m$, we instantiate a group of $K$ independent expert networks $\{\mathcal{K}_m^1, \mathcal{K}_m^2, \dots, \mathcal{K}_m^K\}$ to map the modality embedding $\mathbf{e}m$ into multiple semantic subspaces. This results in a set of multi-view embeddings $\{\mathbf{v}_{e,m}^{(1)}, \mathbf{v}_{e,m}^{(2)}, \dots, \mathbf{v}_{e,m}^{(K)}\}$, each corresponding to a distinct semantic perspective captured by the corresponding expert.

The multi-view design offers two key advantages. First, it encourages representation disentanglement by allowing different experts to specialize in different aspects of the modality, such as appearance, context, or structure. Second, it provides the flexibility to dynamically assess the quality and contribution of each view during fusion, which is especially critical in the presence of noisy or incomplete modality data.

To quantify the information overlap between these views and assess their complementarity, we employ the Mutual Information Neural Estimator (MINE) \cite{limodality}. Each view-specific embedding $\mathbf{v}_{e,m}^{(i)}$ is projected into a probabilistic distribution $P(\mathbf{v}_{e,m}^{(i)})$ using a dedicated MLP $f_m^i$. The mutual information $I(\mathbf{v}_{e,m}^{(i)}; \mathbf{v}_{e,m}^{(j)})$ between any two views then serves as a statistical measure of redundancy: lower mutual information implies that the views encode more distinct, non-overlapping features, and hence are more complementary. Formally, mutual information is defined as:

\begin{equation}
I(X; Y) = \sum_{x \in X} \sum_{y \in Y} p(x, y) \log \frac{p(x, y)}{p(x)p(y)}
\end{equation}
where $X = P(\mathbf{v}_{e,m}^{(i)})$ and $Y = P(\mathbf{v}_{e,m}^{(j)})$ are the learned distributions of two expert views.

To integrate the multi-view information, we propose a mutual information-based adaptive weighting strategy, where the normalized contribution weight $\omega_m^a$ of each view is computed based on a softmax over the negative mutual information with all other views. This approach implicitly promotes diversity-aware feature aggregation, assigning higher weights to views that offer more novel information. The fused intra-modality embedding is thus formulated as:

\begin{equation}
\hat{\mathbf{v}}_{e,m} = \omega_m^a \cdot \mathbf{v}_{e,m} = \sum_{i=1}^{K} 
\left(
\exp\left( - \sum_{j \ne i} I(\mathbf{v}_{e,m}^{(i)};\ \mathbf{v}_{e,m}^{(j)}) \right)
\middle/
\sum_{i'=1}^{K} \exp\left( - \sum_{j \ne i'} I(\mathbf{v}_{e,m}^{(i')};\ \mathbf{v}_{e,m}^{(j)}) \right)
\right)
\cdot \mathbf{v}_{e,m}^{(i)}
\end{equation}

The equation ensures that each view-specific representation $\mathbf{v}_{e,m}^{(i)}$ is weighted according to its uniqueness with respect to others. Specifically, views that share less mutual information with the rest are considered more complementary and thus receive higher weights in the fusion, which is achieved through a softmax-style weighting mechanism, where the negative mutual information $\sum_{j \ne i} I(\mathbf{v}_{e,m}^{(i)};\ \mathbf{v}_{e,m}^{(j)})$ serves as an inverse signal for importance. From an information-theoretic perspective, the strategy effectively promotes diversity-aware representation fusion, favoring views that capture novel or minimally redundant features, while down-weighting overly similar ones. As a result, the aggregated intra-modality embedding $\hat{\mathbf{v}}_{e,m}$ preserves a richer and more discriminative feature composition.

Building upon this, we extend the same principle to inter-modality fusion. After obtaining the intra-modality fused embeddings $\hat{\mathbf{v}}_{e,m}$ for all $m \in \mathcal{M}$, we again compute mutual information $I(\hat{\mathbf{v}}_{e,m}; \hat{\mathbf{v}}_{e,m'})$ between modality pairs. The resulting normalized inter-modality weight $\omega_m^{b}$ reflects the semantic uniqueness of modality $m$ relative to others, allowing the model to prioritize modalities that contribute non-redundant information. The joint entity representation is computed as:
\begin{equation}
\hat{\mathbf{v}}_{e}^{\text{joint}} = \omega_m^{b} \cdot \hat{\mathbf{v}}_{e,m} = \sum_{m \in \mathcal{M}} 
\frac{ 
\exp\left( - \sum_{m' \ne m} I(\hat{\mathbf{v}}_{e,m};\ \hat{\mathbf{v}}_{e,m'}) \right)
}{
\sum_{m'' \in \mathcal{M}} \exp\left( - \sum_{m' \ne m''} I(\hat{\mathbf{v}}_{e,m''};\ \hat{\mathbf{v}}_{e,m'}) \right)
}
\cdot \hat{\mathbf{v}}_{e,m}
\end{equation}

The equation mirrors the intra-modality fusion strategy, where each modality's contribution is determined by its dissimilarity to other modalities. Specifically, modalities that share less mutual information with others are deemed to offer more complementary signals, and thus receive higher softmax-normalized weights. By emphasizing such unique modalities, the model constructs a joint representation $\hat{\mathbf{v}}_{e}^{\text{joint}}$ that integrates semantically diverse and informative features while reducing redundancy across modalities.

The fusion strategy serves two critical purposes: (1) it enables a hierarchical control of information redundancy at both intra- and inter-modality levels, and (2) it provides a principled approach for missing modality compensation, as more informative modalities naturally dominate the final embedding. The complementarity-aware fusion mechanism aligns well with the goal of robust and semantically rich MMKG representation learning, offering theoretical grounding via mutual information theory and practical effectiveness in addressing real-world challenges such as noise and incomplete modality coverage.

\subsection{Entropy-based Negative Sampling}
After obtaining joint multimodal embeddings for entities, we adopt the RotatE score function~\cite{sun2019rotate} to assess the plausibility of a triple $(h, r, t)$:
\begin{algorithm}[H] \footnotesize
\caption{Training procedure of the MoCME framework (Simplified)}
\label{alg:MoCME_simplified}
\begin{algorithmic}[1]
\Require Training set $\mathcal{T}$, modalities $\mathcal{M}$; for each $m\in\mathcal{M}$: pre-trained encoder $PE_m$, projection network $f_m$, expert set $\{\mathcal{K}_m^1, \dots, \mathcal{K}_m^K\}$; negative sampling thresholds $\delta_1,\delta_2$; learning parameters.
\For{each epoch}
    \For{each positive triple $(h,r,t) \in \mathcal{T}$}
        \For{each modality $m \in \mathcal{M}$}
            \State $\mathcal{I}_m(e) \gets$ raw data for $e$ \Comment{for both $h$ and $t$}
            \State $\mathbf{v}_m^r \gets PE_m(\mathcal{I}_m(e))$
            \State $\mathbf{e}_m \gets f_m(\mathbf{v}_m^r)$
            \For{$i=1$ \textbf{to} $K$}
                \State $\mathbf{v}_{e,m}^{(i)} \gets \mathcal{K}_m^i(\mathbf{e}_m)$
            \EndFor
            \State Compute fused modality representation:
            \Statex \quad
            \quad \quad $\hat{\mathbf{v}}_{e,m} \gets \sum_{i=1}^{K}
            \left(
            \exp\left(- \sum_{j \ne i} I\left(\mathbf{v}_{e,m}^{(i)};\ \mathbf{v}_{e,m}^{(j)}\right)\right)
            \middle/
            \sum_{i'=1}^{K} \exp\left(- \sum_{j \ne i'} I\left(\mathbf{v}_{e,m}^{(i')};\ \mathbf{v}_{e,m}^{(j)}\right)\right)
            \right)
            \cdot \mathbf{v}_{e,m}^{(i)}$
        \EndFor

        \For{each modality $m \in \mathcal{M}$}
            \State Compute normalized inter-modality weight:
            \Statex \quad
            \quad \quad $\omega_m^{b} \gets
            \exp\left(- \sum_{m' \ne m} I\left(\hat{\mathbf{v}}_{e,m};\ \hat{\mathbf{v}}_{e,m'}\right)\right) /
            \sum_{m'' \in \mathcal{M}} \exp\left(- \sum_{m' \ne m''} I\left(\hat{\mathbf{v}}_{e,m''};\ \hat{\mathbf{v}}_{e,m'}\right)\right)$
        \EndFor

        \State $\hat{\mathbf{v}}_{e}^{\text{joint}} \gets \sum_{m\in\mathcal{M}} \omega_m^{b} \cdot \hat{\mathbf{v}}_{e,m}$
        \State $\mathcal{S}^+ \gets -\left\|\hat{\mathbf{h}}^{\text{joint}} \circ \mathbf{r} - \hat{\mathbf{t}}^{\text{joint}}\right\|$
        \State Generate negatives $\mathcal{T}'$ by corrupting head or tail.
        \For{each negative triple $(h',r,t') \in \mathcal{T}'$}
            \State $\mathcal{S}^- \gets -\left\|\hat{\mathbf{h'}}^{\text{joint}} \circ \mathbf{r} - \hat{\mathbf{t'}}^{\text{joint}}\right\|$
            \State $p \gets \sigma(\mathcal{S}^-)$
            \State $H \gets - p\log p - (1-p)\log(1-p)$
            \State Set weight $\lambda(h',r,t')$:
            \Statex \quad
            \quad \quad\quad\quad\quad$\lambda(h',r,t') =
            \begin{cases}
            \lambda_{\text{easy}}, & H < \delta_1 \\
            \lambda_{\text{amb}}, & \delta_1 \leq H < \delta_2 \\
            \lambda_{\text{hard}}, & H \geq \delta_2
            \end{cases}$
        \EndFor
        \State $\mathcal{L} \gets -\log\sigma(\mathcal{S}^+) - \sum_{(h',r,t')\in\mathcal{T}'} \lambda(h',r,t')\,\log\sigma(-\mathcal{S}^-)$
        \State Update model parameters via gradient descent.
    \EndFor
\EndFor
\State \textbf{Inference:} For a query $(h,r,?)$ or $(?,r,t)$, compute scores using RotatE, rank candidates, and select the top-$k$.
\end{algorithmic}
\end{algorithm}

\begin{equation}
\mathcal{S}_m(h, r, t) = -\left\| \mathbf{h}_{\text{joint}, m} \circ \mathbf{r}_m - \mathbf{t}_{\text{joint}, m} \right\|
\end{equation}

where $\circ$ denotes the element-wise Hadamard product in the complex space, and $\mathbf{h}_{\text{joint}, m}$, $\mathbf{t}_{\text{joint}, m}$ are the joint embeddings of head and tail entities under modality $m$. A smaller distance indicates higher plausibility.

To construct a meaningful training signal, we generate negative samples by corrupting either the head or tail entity of the positive triples, forming a candidate set $\mathcal{T}'$. However, not all negative samples are equally informative; some are too trivial and offer limited training value, while others may be highly ambiguous or even harder than the ground-truth. To quantify such differences, we compute the prediction probability $p = \sigma(\mathcal{S}_m(h', r, t'))$ for each negative triple $(h', r, t') \in \mathcal{T}'$, and estimate its uncertainty via binary entropy:

\begin{equation}
H(h', r, t') = -p \log p - (1 - p) \log (1 - p)
\end{equation}

The entropy $H$ serves as a natural measure of difficulty and informativeness. Specifically, high-entropy samples correspond to those close to the decision boundary (i.e., prediction confidence $\approx 0.5$), while low-entropy samples are easily separable (i.e., confidence $\rightarrow 0$ or $1$). Based on this, we partition $\mathcal{T}'$ into three categories using two thresholds $\delta_1$ and $\delta_2$ $(\delta_1 < \delta_2)$:

\begin{itemize}
\item \textbf{Easy negatives} ($H < \delta_1$): samples with low entropy and high prediction certainty; often redundant or semantically distant from positives.
\item \textbf{Ambiguous negatives} ($\delta_1 \leq H < \delta_2$): samples with intermediate entropy, often located near the classification boundary.
\item \textbf{Hard negatives} ($H \geq \delta_2$): samples with high entropy, possibly semantically close to true triples and difficult to distinguish.
\end{itemize}

From an optimization perspective, training on easy negatives yields vanishing gradients due to overconfidence, while training exclusively on hard negatives may introduce noise or false negatives. Therefore, we propose a type-aware weighted contrastive loss that balances stability and discriminability. Each negative sample is assigned a scalar weight according to its type:

\begin{equation}
\lambda(h', r, t') =
\begin{cases}
\lambda_{\text{easy}}, & \text{if } H < \delta_1 \\
\lambda_{\text{amb}},  & \text{if } \delta_1 \leq H < \delta_2 \\
\lambda_{\text{hard}}, & \text{if } H \geq \delta_2 \\
\end{cases}
\quad \text{with } \lambda_{\text{easy}} < \lambda_{\text{amb}} < \lambda_{\text{hard}}
\end{equation}

The final loss function is formulated as:

\begin{equation}
\mathcal{L}_{\text{final}} = -\log \sigma(\mathcal{S}_m^+) - \sum_{(h', r, t') \in \mathcal{T}'} \lambda(h', r, t') \cdot \log \sigma(-\mathcal{S}_m^-)
\end{equation}
The equation explicitly aligns with the principle of \textit{curriculum learning}, where the model learns from easier examples before focusing on more ambiguous and difficult ones. Moreover, it avoids excessive gradient saturation and overfitting to trivial negatives while leveraging hard samples as valuable contrastive signals. The entropy-based weighting also provides a continuous and theoretically grounded metric for sample importance, differentiating our method from existing approaches that rely on random or frequency-based weighting schemes.

During inference, the MMKGC model performs standard link prediction\cite{bordes2013translating} by ranking candidate triples based on their RotatE scores. Given a query $(h, r, ?)$ or $(?, r, t)$, the score $\mathcal{S}_m(h, r, e)$ or $\mathcal{S}_m(e, r, t)$ is computed for all candidate entities $e \in E$. Entities are then ranked by score in descending order, and the top-$k$ candidates are selected as predictions.

\section{Experiments}
\begin{table*}[ht] \footnotesize
\centering 
\caption{Statistics of the five multimodal knowledge graphs (MMKGs) in the WildKGC benchmark. The table includes the number of entities, relations, and triples in each dataset, as well as the number of entities with available modality information (image, text, numeric, audio, and video).}

\setlength{\tabcolsep}{4pt} 
\begin{tabular}{l|ccccc|ccccc}
\toprule
\multirow{2}{*}{Dataset} &  \multirow{2}{*}{\#Entity} &  \multirow{2}{*}{\#Relation} &  \multirow{2}{*}{\#Train} &  \multirow{2}{*}{\#Valid} &  \multirow{2}{*}{\#Test} & Image & Text & Numeric & Audio & Video \\
\cmidrule(lr){7-11} 
 & & & & & & \multicolumn{5}{c}{Num} \\
\midrule
MKG-W [57] & 15000 & 169 & 34196 & 4276 & 4274 & 14463 & 14123 & - & - & - \\
MKG-Y [57] & 15000 & 28 & 21310 & 2665 & 2663 & 14244 & 12305 & - & - & - \\
DB15K [32] & 12842 & 279 & 79222 & 9902 & 9904 & 12818 & 9078 & 11022 & - & - \\
TIVA [52] & 11858 & 16 & 20071 & 2000 & 2000 & 11636 & 11858 & - & 2441 & 10269 \\
KVC16K [34] & 16015 & 4 & 180190 & 22523 & 22525 & 14822 & 14822 & - & 14822 & 14822 \\
\bottomrule
\end{tabular}
\label{data}
\end{table*}

\subsection{Implementation Details}

We conduct all experiments on a Linux server equipped with the Ubuntu 20.04.1 LTS operating system, an Intel Xeon Gold 6226R CPU (2.90GHz), and a single NVIDIA RTX 4090 GPU with 24GB memory. All methods are implemented in PyTorch 1.13 with CUDA 11.7 and Python 3.9. For the training of our proposed MoCME framework, we largely follow the hyperparameter configuration and baseline setup reported in NATIVE~\cite{zhang2024native}. Specifically, we adopt the Adam optimizer~\cite{diederik2014adam} with a learning rate of $1 \times 10^{-4}$. The batch size is set to 1024, and the training process runs for 1000 epochs with early stopping based on validation MRR. All embeddings, including entities, relations, and modality-specific projections, are initialized with Xavier uniform initialization and set to a fixed dimension of 256, consistent with~\cite{zhang2024native}.

For the information entropy-based negative sampling mechanism, we empirically set the entropy thresholds $\delta_1 = 0.2$ and $\delta_2 = 0.8$ to divide the negative sample space into three difficulty levels. The corresponding loss weights for different negative types are set to $\lambda_{\text{easy}} = 0.5$, $\lambda_{\text{amb}} = 1.5$, and $\lambda_{\text{hard}} = 1.2$, respectively. We also ablate this setting in Section \ref{delta}. All reported results are averaged over three independent runs with different random seeds. The best checkpoint on the validation set is selected for final testing. For a fair comparison, we retrain all baselines under identical settings when the original code or checkpoints are available.

\subsection{Datasets}
To better demonstrate the effectiveness and robustness of MoCME under diverse multimodal and structural configurations, we conduct extensive experiments on five widely-used MMKGC benchmarks, each characterized by distinct modality combinations, coverage densities, and relational complexities. Detailed statistics are provided in Table~\ref{data}.

\begin{itemize}
    \item \textbf{MKG-Y} and \textbf{MKG-W} \cite{xu2022relation} are two multimodal knowledge graphs constructed from subsets of YAGO and Wikidata, respectively. Both datasets contain only \textit{visual} and \textit{textual} modalities, providing a clean yet challenging testbed to evaluate modality fusion and modality complementarity in the absence of structural complexity or numerical information. Specifically, MKG-Y and MKG-W consist of 15,000 entities each. MKG-Y includes 14,244 entities with image data and 12,305 with text data, while MKG-W provides 14,463 entities with images and 14,123 with texts. These datasets are particularly suitable for analyzing vision-text interactions and baseline modality alignment performance.
    
    \item \textbf{DB15K} \cite{liu2019mmkg} is a commonly adopted MMKG derived from DBpedia and contains entity-level numerical attributes in addition to textual descriptions and images. The dataset consists of 12,842 entities, 279 relations, and 79,222 training triples. Among these entities, 12,818 have image data, 9,078 have text, and 11,022 are associated with numeric information. The introduction of the numeric modality increases the heterogeneity of the data and offers new challenges in modality integration due to differences in data type, scale, and semantic granularity. This dataset allows us to evaluate the model's ability to adapt to non-sequential, low-dimensional features in a multimodal context.
    
    \item \textbf{TIVA} \cite{wang2023tiva} is a recently proposed MMKG that includes four distinct modalities: text, image, audio, and video. It contains 11,858 entities, with 11,636 entities having image data, 11,858 with text, 2,441 with audio, and 10,269 with video. Notably, it contains a richer set of video information, reflecting real-world multimedia settings where temporal and acoustic semantics are crucial. The diversity of modalities in TIVA enables the assessment of MoCME under high-modality-complexity conditions and supports the exploration of inter-modality complementarity across heterogeneous signal types.
    
    \item \textbf{KVC16K} \cite{pan2022kuaipedia} is a large-scale MMKG collected from Kuaipedia, encompassing the same four modalities as TIVA. It consists of 16,015 entities, and notably, 14,822 entities are associated with each of the four modalities: image, text, audio, and video. It includes denser audio modality coverage compared to TIVA, making it particularly valuable for examining the impact of modality imbalance on MMKGC tasks. KVC16K serves as a complementary benchmark to TIVA, offering insights into how well models can exploit audio-dominant contexts when other modalities are sparse or missing.
\end{itemize}

These datasets provide a comprehensive empirical foundation for evaluating MoCME under varying degrees of modality richness, information imbalance, and semantic granularity, which are key to understanding the real-world applicability of multimodal knowledge graph reasoning systems.


\subsection{Baseline Methods}
We compare the performance of the proposed framework with 19 different state-of-the-art (SOTA) methods, which are divided into three categories: traditional KGC, multi-modal KGC, negative sampling methods, to conduct a detailed and comprehensive comparative analysis.
 
\subsubsection{Traditional KGC methods} Five SOTA conventional KGC methods are selected for performance comparison, including TransE \cite{bordes2013translating}, DistMult \cite{yang2014embedding}, ComplEx \cite{trouillon2016complex}, RotatE \cite{sun2019rotate}, and PairRE \cite{chao2020pairre}, which learn structural embeddings optimized by different elegant score functions from a given KG without utilizing any multimodal information.

\subsubsection{Multi-modal KGC methods} To ensure a fairer performance comparison, we select 10 distinct SOTA MMKGC methods for evaluation that integrate both multimodal information or structural knowledge of triples, including IKRL \cite{xie2016image}, TBKGC \cite{mousselly2018multimodal}, TransAE \cite{wang2019multimodal}, RSME \cite{wang2021visual}, MMKRL \cite{lu2022mmkrl}, VBKGC \cite{zhang2022knowledge}, OTKGE \cite{cao2022otkge}, IMF \cite{li2023imf}, QEB \cite{wang2023tiva}, VISTA \cite{lee2023vista}.

\subsubsection{Negative sampling Methods} Some latest negative sampling methods including KBGAN \cite{cai2017kbgan}, MANS \cite{zhang2023modality}, MMRNS \cite{xu2022relation} and NATIVE \cite{zhang2024native} are utilized for performance comparison. KBGAN employs adversarial sampling with reinforcement learning for traditional KGC, while MANS, MMRNS and NATIVE are designed for MMKGC, leveraging multimodal data to enhance negative sampling quality.


\begin{table*}[h!]\footnotesize
\centering
\renewcommand{\arraystretch}{1.2}
\setlength{\tabcolsep}{2pt} 
\caption{Performance comparison across datasets and methods}
\label{tab:results}
\begin{tabular}{c|c|cc|cc|ccc|ccc|ccc}
\hline
\multirow{2}{*}{Method} & \multirow{2}{*}{Modality} & \multicolumn{2}{c|}{MKG-W} & \multicolumn{2}{c|}{MKG-Y} & \multicolumn{3}{c|}{DB15K} & \multicolumn{3}{c|}{KVC16k} & \multicolumn{3}{c}{TIVA} \\ 
 &  & MRR & Hit@1 & MRR & Hit@1 & MRR & Hit@1 & Hit@10 & MRR & Hit@1 & Hit@10 & MRR & Hit@1 & Hit@10 \\ \hline
\multicolumn{15}{c}{\textit{Uni-modal KGC Methods}} \\ \hline
TransE & S & 29.19 & 21.06 & 30.73 & 23.45 & 24.86 & 12.78 & 47.07 & 8.54 & 0.64 & 23.42 & 83.85 & 83.20 & 84.15 \\ 
DistMult & S & 29.09 & 15.93 & 25.04 & 19.33 & 23.03 & 14.78 & 39.59 & 6.57 & 3.03 & 12.61 & 82.27 & 81.15 & 84.22 \\ 
ComplEx & S & 24.93 & 19.09 & 28.71 & 22.26 & 27.48 & 18.37 & 45.37 & 12.85 & 7.48 & 23.18 & 80.67 & 77.67 & 86.10 \\ 
RotatE & S & 33.67 & 26.80 & 34.95 & 29.10 & 29.28 & 17.87 & 49.66 & 14.33 & 8.25 & 26.17 & 84.59 & 83.47 & 86.95 \\ 
PairRE & S & 34.40 & 28.24 & 32.01 & 25.53 & 31.13 & 21.62 & 49.30 & - & - & - & - & - & - \\ 
\hline
\multicolumn{15}{c}{\textit{Multi-modal KGC Methods}} \\ \hline
IKRL & S+I & 32.36 & 26.11 & 33.22 & 30.37 & 26.82 & 14.09 & 49.09 & 11.11 & 5.42 & 22.39 & 67.71 & 63.72 & 75.67 \\ 
TBKGC & S+T & 31.48 & 25.31 & 33.99 & 30.47 & 28.40 & 15.61 & 49.86 & 5.39 & 0.35 & 15.52 & 81.57 & 78.75 & 86.05 \\ 
TransAE & S+T & 30.00 & 21.23 & 28.10 & 25.31 & 28.09 & 21.25 & 41.17 & 10.81 & 5.31 & 21.89 & 79.57 & 74.95 & 88.07 \\ 
MMKRL$^\dagger$ & S+T & 30.10 & 22.16 & 36.81 & 31.66 & 26.81 & 13.85 & 49.39 & 8.78 & 3.89 & 18.34 & 85.03 & 81.92 & 90.10 \\ 
RSME & S+I & 29.23 & 23.36 & 34.44 & 31.78 & 29.76 & 24.15 & 40.29 & 12.31 & 7.14 & 22.05 & 40.01 & 30.55 & 51.35 \\ 
VBKGC & S+I+T & 30.61 & 24.91 & 37.04 & 33.76 & 30.61 & 19.75 & 49.44 & 14.66 & 8.28 & 27.04 & 74.07 & 66.87 & 85.85 \\ 
OTKGE & S+I+T & 34.36 & 28.85 & 35.51 & 31.97 & 23.86 & 18.45 & 34.23 & 8.77 & 5.01 & 15.55 & 35.28 & 30.45 & 41.98 \\ 
IMF & S+I+T & 34.50 & 28.77 & 35.79 & 32.95 & 32.25 & 24.20 & 48.19 & 12.01 & 7.42 & 21.01 & 55.46 & 41.87 & 77.57 \\ 
QEB & All & 32.38 & 25.47 & 34.37 & 29.49 & 28.18 & 14.82 & 51.55 & 12.06 & 5.57 & 25.01 & 74.25 & 66.10 & 88.35 \\ 
VISTA & S+I+T & 32.91 & 26.12 & 30.45 & 24.87 & 30.42 & 22.49 & 45.94 & 11.89 & 6.97 & 21.27 & 76.07 & 70.67 & 86.60 \\ 
\hline
\multicolumn{15}{c}{\textit{Negative Sampling Methods}} \\ \hline
KBGAN$^\dagger$ & S & 29.47 & 22.21 & 29.71 & 22.81 & 25.73 & 9.91 & 51.93 & 13.72 & 7.54 & 25.88 & 85.44 & 82.45 & 90.10 \\ 
MANS & S+I & 30.88 & 24.89 & 29.03 & 25.25 & 28.82 & 16.87 & 49.26 & 10.42 & 5.21 & 20.45 & 85.70 & 82.70 & 90.62 \\ 
MMRNS & S+I+T & 35.03 & 28.59 & 35.93 & 30.53 & 32.68 & 23.01 & 51.01 & 13.31 & 7.51 & 24.68 & 83.12 & 83.05 & 83.25 \\ 
NativE & All & 36.58 & 29.56 & 39.04 & 34.79 & 37.16 & 28.01 & 54.13 & 15.76 & 9.23 & 28.55 & 92.10 & 91.40 & 92.85 \\ \hline
MoCME & All & \textbf{37.79} & \textbf{30.81} & \textbf{40.37} & \textbf{36.21} & \textbf{39.62} & \textbf{29.71} & \textbf{55.36} & \textbf{18.32} & \textbf{11.23} & \textbf{30.76} & \textbf{94.58} & \textbf{93.32} & \textbf{95.61} \\ 
\hline
\end{tabular}
\label{main}
\end{table*}

\subsection{Main Results}
Table~\ref{main} presents a comprehensive comparison of MoCME with representative state-of-the-art methods across five MMKGC datasets and three commonly used evaluation metrics: MRR, Hit@1, and Hit@10. As observed, MoCME consistently achieves the best overall performance on all datasets, demonstrating its superior capability in modeling and integrating complementary multimodal information.

On \textbf{MKG-W} and \textbf{MKG-Y}, which contain only image and text modalities, MoCME shows modest improvements compared to the strongest baseline (NativE), achieving MRR gains of \textbf{+1.21\%} and \textbf{+1.33\%}, respectively. The relatively limited improvement can be attributed to the absence of additional modalities, which constrains the full expressiveness of our modality complementarity-aware fusion mechanism. Nevertheless, MoCME still achieves the highest Hit@1 and Hit@10 on both datasets, indicating more accurate top-ranked predictions.

More significant gains are observed on datasets with richer multimodal inputs. On \textbf{DB15K}, which includes numeric information in addition to visual and textual modalities, MoCME surpasses all baselines with a \textbf{+2.46\%} increase in MRR over NativE, along with clear gains in Hit@1 (\textbf{+1.70\%}) and Hit@10 (\textbf{+1.23\%}). This highlights the model's ability to incorporate low-dimensional, non-sequential features effectively.

The advantage of MoCME becomes even more prominent on \textbf{KVC16K} and \textbf{TIVA}, both of which include all four modality types: image, text, audio, and video. On KVC16K, MoCME outperforms NativE by \textbf{+2.56\%} MRR, \textbf{+1.67\%} Hit@1, and \textbf{+2.21\%} Hit@10, demonstrating strong performance in environments with highly imbalanced modality distributions. Most notably, on \textbf{TIVA}, MoCME achieves the best performance across all metrics, with an MRR of \textbf{94.58}, Hit@1 of \textbf{93.32}, and Hit@10 of \textbf{95.61}. These results correspond to relative gains of \textbf{+2.48\%}, \textbf{+1.92\%}, and \textbf{+2.76\%} over the previous best method.

In summary, the empirical results demonstrate that while MoCME maintains competitive performance under low-modality settings, its strength becomes increasingly evident as modality complexity grows. The model’s ability to adaptively fuse and emphasize complementary modality features—especially under heterogeneous and noisy inputs—contributes to its consistent outperformance across all evaluation scenarios.

\begin{table}[h!]
\centering
\renewcommand{\arraystretch}{1.2}
\setlength{\tabcolsep}{6pt}
\caption{Ablation study of modality and module components in MoCME on the \textbf{DB15K} dataset. Each row shows which components are enabled (\checkmark) or disabled (\xmark), along with the resulting performance.}
\label{tab:ablation_grid}
\begin{tabular}{cccccc|ccc}
\hline
\multicolumn{6}{c|}{Modality and Components} & \multicolumn{3}{c}{Metrics}  \\ 
\hline
Image & Text & Numeric & $\omega_m^a$ & $\omega_m^b$ & $\mathcal{L}_{gc}$ & MRR & Hit@1 & Hit@10 \\ 
\hline
\xmark & \checkmark & \checkmark & \checkmark & \checkmark & \checkmark &  39.05 & 29.46 & 55.07 \\
\checkmark & \xmark & \checkmark & \checkmark & \checkmark & \checkmark &  38.41 & 28.75 & 54.62 \\
\checkmark & \checkmark & \xmark & \checkmark & \checkmark & \checkmark &  38.89 & 29.31 & 54.94 \\
\checkmark & \checkmark & \checkmark & \xmark & \checkmark & \checkmark &  38.76 & 29.14 & 54.86 \\
\checkmark & \checkmark & \checkmark & \checkmark & \xmark & \checkmark &  37.64 & 28.55 & 53.42 \\
\checkmark & \checkmark & \checkmark & \checkmark & \checkmark & \xmark &  38.12 & 28.73 & 54.37 \\
\checkmark & \checkmark & \checkmark & \checkmark & \checkmark & \checkmark  & \textbf{39.62} & \textbf{29.71} & \textbf{55.36} \\
\hline
\end{tabular}
\label{component}
\end{table}


\subsection{Ablation Study}
To validate the effectiveness of the proposed MoCME framework, we conduct extensive ablation studies, including experiments on individual components, modalities, fusion strategies, and backbone choices for processing raw features. The results are presented in Table \ref{component}, \ref{weight}, and \ref{encoder}.

\subsubsection{Ablations on Component and Modality}
The component and modality ablation experiments (Table~\ref{component}) provide comprehensive insights into the contribution of each modality and fusion component within the MoCME framework on the \textbf{DB15K} dataset. From the modality perspective, removing any of the three modalities (image, text, or numeric) leads to a noticeable drop in performance, confirming that each modality provides complementary semantic cues that benefit final prediction. Among them, the absence of image features causes the largest decrease in MRR ($\textbf{-0.57}$), followed by text ($\textbf{-1.21}$), and numeric ($\textbf{-0.73}$), indicating that all three modalities play indispensable roles.

From the architectural perspective, removing the inter-modality complementarity module ($\omega_m^b$) leads to the most significant performance degradation, with MRR dropping from \textbf{39.62} to \textbf{37.64}, suggesting that modeling cross-modal interactions is more critical than intra-modal fusion alone. The intra-modality complementarity module ($\omega_m^a$) also contributes meaningfully, albeit to a lesser extent ($\textbf{-0.86}$ MRR). In addition, removing the entropy-weighted contrastive loss ($\mathcal{L}_{gc}$) results in a performance decline ($\textbf{-1.50}$ Hit@10), underscoring its importance in enforcing discriminative representation learning across modalities.

Overall, these results confirm that both modality-aware fusion and complementary feature modeling are key to the effectiveness of MoCME, with inter-modality reasoning and contrastive learning playing especially pivotal roles.

\subsubsection{Ablations on fusion strategies}
As shown in Table~\ref{weight}, we compare the proposed complementarity-based fusion with three commonly used alternatives: concatenation, element-wise product, and gating. The results clearly demonstrate that simple fusion strategies fall short in capturing the nuanced interactions between multi-view features. Specifically, using concatenation results in a dramatic performance degradation (MRR drops to \textbf{22.34}), suggesting that such naive combination fails to preserve informative dependencies across modalities or expert views. While attention-based fusion methods are often used in related tasks to learn importance weights, we found them impractical in our setting due to the prohibitive memory overhead—particularly considering the large number of entities and multi-modal views in the MMKG, which make full pairwise attention computation infeasible. Element-wise product performs better (MRR = \textbf{31.96}), but still lags behind due to its limited expressiveness and inability to adaptively weigh feature importance.

\subsubsection{Ablations on backbone architectures}
Table \ref{encoder} presents the comparison of different backbone architectures used to project and transform modality features into shared embedding spaces. Replacing the Complementary Mixture of Experts (CMoE) with a simple multi-layer perceptron (MLP) results in an MRR drop from \textbf{39.62} to \textbf{37.63}, while a linear projection performs even worse (MRR = \textbf{36.84}). This performance degradation confirms that shallow or static architectures are insufficient to model the complex, view-specific dynamics required in multi-modal settings. In contrast, the proposed CMoE architecture allows for modular expert specialization and adaptive feature routing, guided by mutual information–driven complementarity. It enables the model to disentangle redundant and complementary patterns across views, leading to stronger representation learning. These results underscore the importance of designing an expressive and flexible backbone in multimodal knowledge graph completion tasks.

\subsubsection{Ablations on the number of experts}
\begin{figure}
    \centering
    \includegraphics[scale=0.5]{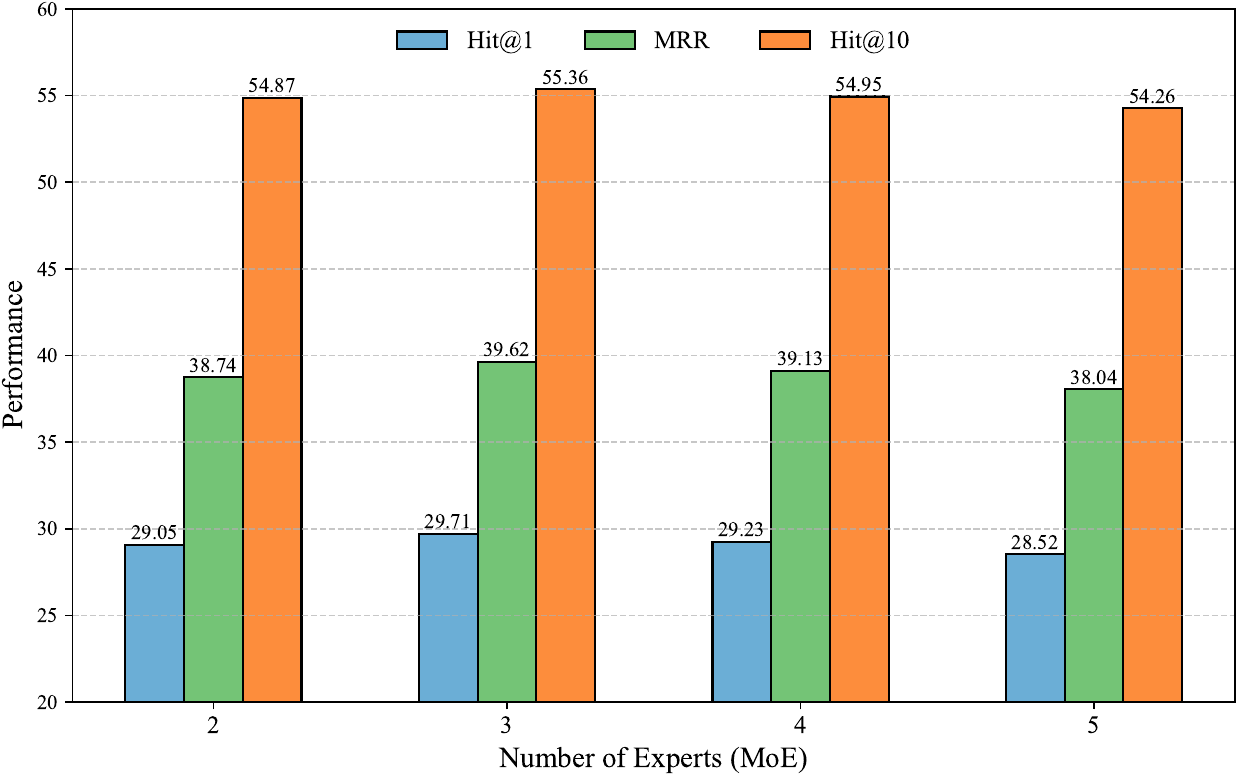}
    \caption{Ablation study on the number of experts in MoCME framework}
    \label{per}
\end{figure}

The ablation study in Figure~\ref{per} highlights the importance of selecting an appropriate number of experts for optimal performance. For the DB15K dataset, which includes three modalities (text, image, and structure), the best performance is achieved when the number of experts is set to 3, matching the number of modalities. This alignment supports our design intuition that each expert can specialize in one modality's representation space, thereby facilitating more effective intra-modality fusion. As the number of experts increases beyond the number of modalities (e.g., 4 or 5), performance slightly degrades across all metrics (Hit@1, MRR, and Hit@10), likely due to over-fragmentation of the embedding space and increased difficulty in aggregation. Conversely, with fewer experts (e.g., 2), the model underperforms, suggesting insufficient modeling capacity to capture modality-specific variations. Interestingly, we observe a consistent performance peak around the modality count across all three datasets, indicating that the number of experts should ideally reflect the number of available modalities. This further validates the complementarity-aware expert design in MoCME, and provides empirical guidance for setting the number of experts in multimodal knowledge graph tasks.

\begin{table}[h!]
\centering
\renewcommand{\arraystretch}{1.2}
\setlength{\tabcolsep}{10pt} 
\caption{Performance comparison of different fusion strategies.}
\label{tab:ablation}
\begin{tabular}{l|ccc}
\hline
\textbf{Setting} & \textbf{MRR} & \textbf{Hit@1} & \textbf{Hit@10} \\ \hline
w/ concat & 22.34 & 9.57 & 50.21 \\ 
w/ product & 31.96 & 23.42 & 51.44 \\ 
w/ gate & 38.05 & 28.76 & 54.09 \\ \hline
\textbf{MoCME} & \textbf{39.62} & \textbf{29.71} & \textbf{55.36} \\ \hline
\end{tabular}
\label{weight}
\end{table}

\begin{table}[h!]
\centering
\renewcommand{\arraystretch}{1.2}
\setlength{\tabcolsep}{10pt} 
\caption{Performance evaluation of different backbone architectures.}
\label{tab:ablation}
\begin{tabular}{l|ccc}
\hline
\textbf{Setting} & \textbf{MRR} & \textbf{Hit@1} & \textbf{Hit@10} \\ \hline
w/ MLP & 37.63 & 28.24 & 54.31 \\ 
w/ Linear & 36.84 & 27.69 & 53.65 \\ 
w/ CMoE & \textbf{39.62} & \textbf{29.71} & \textbf{55.36} \\ \hline
\end{tabular}
\label{encoder}
\end{table}

\subsubsection{Ablations on $\delta_1$ and $\delta_2$} \label{delta}
To investigate the effect of the entropy thresholds in our negative sampling strategy, we conduct a grid search over $\delta_1 \in {0.1, 0.2, 0.3, 0.4, 0.5}$ and $\delta_2 \in {0.5, 0.6, 0.7, 0.8, 0.9}$, and report the MRR performance of different combinations in Figure~\ref{per}. The results reveal several key insights. First, a proper margin between the two thresholds is essential. When $\delta_1=0.2$ and $\delta_2=0.8$, the model achieves the best MRR of \textbf{39.62}, indicating that an effective separation of easy, ambiguous, and hard negatives helps optimize contrastive learning. Second, we observe that setting $\delta_1$ too low (e.g., 0.1) leads to degraded performance across all $\delta_2$, likely due to oversampling of trivial negatives. On the other hand, increasing $\delta_1$ beyond 0.3 also causes performance decline, showing that overly conservative classification of negatives may reduce training diversity. Third, as $\delta_1$ increases row-wise, the performance drops significantly, especially when $\delta_1 = 0.4$ or $0.5$, regardless of $\delta_2$. This trend confirms that maintaining a balanced proportion of ambiguous samples is crucial. Furthermore, we also observe that increasing $\delta_2$ (i.e., expanding the range of ambiguous samples while reducing the scope of hard negatives) generally improves performance when $\delta_1$ is well-chosen. This trend highlights the benefit of incorporating a moderate amount of hard negatives into training. Hard samples—those close to the positive decision boundary—help the model refine fine-grained distinctions and prevent overfitting to easy or ambiguous cases. However, when $\delta_2$ is set too high (e.g., 0.9), the proportion of hard negatives becomes too small, limiting their contribution. Conversely, setting $\delta_2$ too low (e.g., 0.5) overly emphasizes hard samples and may introduce noise or false negatives, leading to unstable optimization. Overall, the experimental results show that the entropy-based partitioning mechanism is highly sensitive to threshold settings. A well-calibrated balance between ambiguous and hard samples, achieved via appropriate $\delta_1$ and $\delta_2$, is key to maximizing contrastive learning performance

\begin{figure}
    \centering
    \includegraphics[scale=0.6]{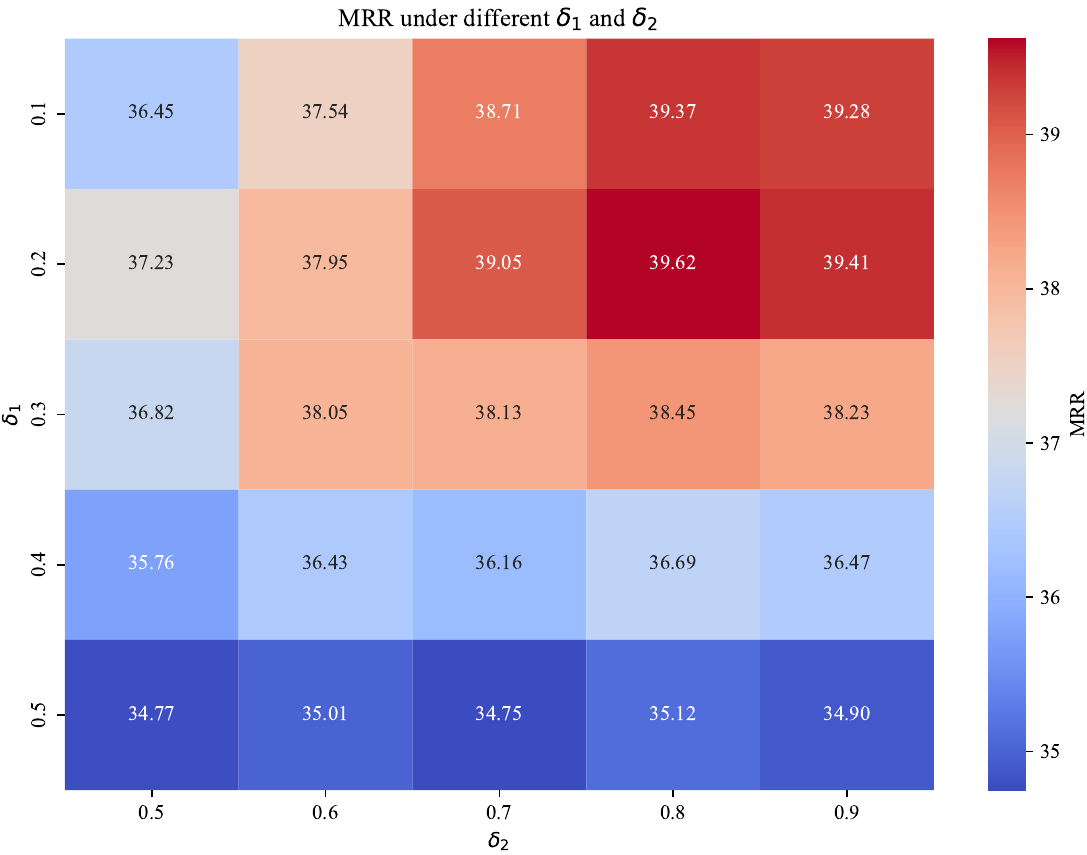}
    \caption{MRR performance under different threshold settings of $\delta_1$ and $\delta_2$ in the entropy-based negative sampling strategy. $\delta_1$ and $\delta_2$ control the partitioning of negative samples into easy, ambiguous, and hard types. The result shows that setting $\delta_1=0.2$ and $\delta_2=0.8$ achieves the best performance.}
    \label{mrr}
\end{figure}

\subsubsection{Time complexity}
In terms of time complexity, traditional KGE methods incur a complexity of \(O(Nd)\), where \(N\) is the number of entities and \(d\) is the embedding dimension, since they only process the graph’s structural information. In contrast, attention-based multimodal fusion methods extend this by integrating cross-modal interactions via attention mechanisms, resulting in a complexity of \(O(MNd^2)\), where \(M\) denotes the number of modalities --- this is due to the quadratic dependency on \(d\) when computing attention weights between all pairs of feature components. Our proposed MoCME framework, however, incorporates multiple expert networks and complementarity calculations. Specifically, for each modality \(m\), \(K\) expert networks produce multi-view embeddings, and the subsequent intra-modality fusion involves aggregating these views with weights computed from their pairwise mutual information. This step roughly incurs an additional cost of \(O(K)\) per modality per entity, leading to an overall cost of approximately \(O(MKN)\). Although we also compute mutual information for complementary weighting, the cost of these computations is generally bounded by a constant when \(K\) and \(M\) remain small.

Therefore, while MoCME introduces extra overhead compared to traditional KGE (\(O(Nd)\)), its overall complexity \(O(MKN)\) is considerably lower than that of attention-based methods \(O(MNd^2)\) in practical settings where \(d\) (the embedding dimension) is typically much larger than the product \(MK\). The proposed MoCME framework provides a favorable trade-off between computational efficiency and the enhanced representational power achieved by effectively modeling both intra- and inter-modality complementarity.

\section{Conclusion}
In this paper, we introduce the Mixture of Complementary Modality Experts (MoCME) framework for the task of Multimodal Knowledge Graph Completion (MMKGC), aiming to uncover hidden world knowledge in multimodal knowledge graphs by fully exploiting both structural and multimodal entity information. Unlike existing MMKGC methods that rely on attention or gate-based fusion while neglecting the inherent complementarity among modalities, our approach explicitly models and leverages both intra-modal and inter-modal complementarity to improve the expressiveness and robustness of entity representations. To this end, we design a Complementarity-guided Modality Knowledge Fusion (CMKF) module that effectively integrates multi-view and multi-modal embeddings, and introduce an Entropy-guided Negative Sampling (EGNS) mechanism that dynamically emphasizes informative and uncertain negative samples. By incorporating an entropy-weighted contrastive loss, MoCME improves training efficiency and enhances model generalization. Extensive experiments on five benchmark datasets demonstrate that MoCME consistently outperforms existing state-of-the-art methods, achieving significant gains across various evaluation metrics. This work highlights the importance of complementarity-aware fusion in MMKGC and offers a robust solution capable of handling missing, incomplete, or noisy modalities in real-world scenarios.

Looking ahead, there are several exciting directions for future work. First, we aim to extend the MoCME framework to handle even more complex multimodal settings, such as incorporating temporal or spatial information, which further enhances the model's ability to capture dynamic knowledge. Additionally, exploring the integration of more advanced techniques for modality alignment, such as unsupervised or semi-supervised learning, is capable of improving performance in scenarios with limited labeled data. Finally, we plan to investigate the scalability of MoCME to large-scale knowledge graphs, enabling its application in real-world, industrial-level knowledge graph completion tasks.


\subsubsection*{Acknowledgments}
This work is supported by This work is supported by National Key R\&D Program of China (2023YFC3502900), National Natural Science Foundation of China
(granted No. 62192731) and National Key R\&D Program of China
(2021YFF1201100). And the authors greatly appreciate the anonymous reviewers' suggestions and the editor's encouragement.

\bibliographystyle{elsarticle-num}
\bibliography{cite}

\begin{thebibliography}{10}
\expandafter\ifx\csname url\endcsname\relax
  \def\url#1{\texttt{#1}}\fi
\expandafter\ifx\csname urlprefix\endcsname\relax\def\urlprefix{URL }\fi
\expandafter\ifx\csname href\endcsname\relax
  \def\href#1#2{#2} \def\path#1{#1}\fi

\bibitem{bordes2013translating}
A.~Bordes, N.~Usunier, A.~Garcia-Duran, J.~Weston, O.~Yakhnenko, Translating
  embeddings for modeling multi-relational data, Advances in neural information
  processing systems 26 (2013).

\bibitem{chao2020pairre}
L.~Chao, J.~He, T.~Wang, W.~Chu, Pairre: Knowledge graph embeddings via paired
  relation vectors, arXiv preprint arXiv:2011.03798 (2020).

\bibitem{sun2019rotate}
Z.~Sun, Z.-H. Deng, J.-Y. Nie, J.~Tang, Rotate: Knowledge graph embedding by
  relational rotation in complex space, arXiv preprint arXiv:1902.10197 (2019).

\bibitem{trouillon2016complex}
T.~Trouillon, J.~Welbl, S.~Riedel, {\'E}.~Gaussier, G.~Bouchard, Complex
  embeddings for simple link prediction, in: International conference on
  machine learning, PMLR, 2016, pp. 2071--2080.

\bibitem{hu2024knowledge}
J.~Hu, H.~Yang, F.~Teng, S.~Du, T.~Li, A knowledge graph completion model based
  on triple level interaction and contrastive learning, Pattern Recognition 156
  (2024) 110783.

\bibitem{wang2022entity}
Z.~Wang, L.~Yang, Z.~Lei, A.~U. Haq, D.~Zhang, S.~Yang, A.~O. Francis, An
  entity-weights-based convolutional neural network for large-sale complex
  knowledge embedding, Pattern Recognition 131 (2022) 108841.

\bibitem{zheng2021knowledge}
W.~Zheng, L.~Yin, X.~Chen, Z.~Ma, S.~Liu, B.~Yang, Knowledge base graph
  embedding module design for visual question answering model, Pattern
  recognition 120 (2021) 108153.

\bibitem{DBLP:conf/aaai/ShangTHBHZ19}
C.~Shang, Y.~Tang, J.~Huang, J.~Bi, X.~He, B.~Zhou,
  \href{https://doi.org/10.1609/aaai.v33i01.33013060}{End-to-end
  structure-aware convolutional networks for knowledge base completion}, in:
  The Thirty-Third {AAAI} Conference on Artificial Intelligence, {AAAI} 2019,
  The Thirty-First Innovative Applications of Artificial Intelligence
  Conference, {IAAI} 2019, The Ninth {AAAI} Symposium on Educational Advances
  in Artificial Intelligence, {EAAI} 2019, Honolulu, Hawaii, USA, January 27 -
  February 1, 2019, {AAAI} Press, 2019, pp. 3060--3067.
\newblock \href {https://doi.org/10.1609/AAAI.V33I01.33013060}
  {\path{doi:10.1609/AAAI.V33I01.33013060}}.
\newline\urlprefix\url{https://doi.org/10.1609/aaai.v33i01.33013060}

\bibitem{DBLP:conf/naacl/NguyenNNP18}
D.~Q. Nguyen, T.~D. Nguyen, D.~Q. Nguyen, D.~Q. Phung,
  \href{https://doi.org/10.18653/v1/n18-2053}{A novel embedding model for
  knowledge base completion based on convolutional neural network}, in: M.~A.
  Walker, H.~Ji, A.~Stent (Eds.), Proceedings of the 2018 Conference of the
  North American Chapter of the Association for Computational Linguistics:
  Human Language Technologies, NAACL-HLT, New Orleans, Louisiana, USA, June
  1-6, 2018, Volume 2 (Short Papers), Association for Computational
  Linguistics, 2018, pp. 327--333.
\newblock \href {https://doi.org/10.18653/V1/N18-2053}
  {\path{doi:10.18653/V1/N18-2053}}.
\newline\urlprefix\url{https://doi.org/10.18653/v1/n18-2053}

\bibitem{DBLP:conf/iclr/VashishthSNT20}
S.~Vashishth, S.~Sanyal, V.~Nitin, P.~P. Talukdar,
  \href{https://openreview.net/forum?id=BylA\_C4tPr}{Composition-based
  multi-relational graph convolutional networks}, in: 8th International
  Conference on Learning Representations, {ICLR} 2020, Addis Ababa, Ethiopia,
  April 26-30, 2020, OpenReview.net, 2020.
\newline\urlprefix\url{https://openreview.net/forum?id=BylA\_C4tPr}

\bibitem{cai2017kbgan}
L.~Cai, W.~Y. Wang, Kbgan: Adversarial learning for knowledge graph embeddings,
  arXiv preprint arXiv:1711.04071 (2017).

\bibitem{zhang2023modality}
Y.~Zhang, M.~Chen, W.~Zhang, Modality-aware negative sampling for multi-modal
  knowledge graph embedding, in: 2023 International Joint Conference on Neural
  Networks (IJCNN), IEEE, 2023, pp. 1--8.

\bibitem{xu2022relation}
D.~Xu, T.~Xu, S.~Wu, J.~Zhou, E.~Chen, Relation-enhanced negative sampling for
  multimodal knowledge graph completion, in: Proceedings of the 30th ACM
  international conference on multimedia, 2022, pp. 3857--3866.

\bibitem{zhang2022knowledge}
Y.~Zhang, W.~Zhang, Knowledge graph completion with pre-trained multimodal
  transformer and twins negative sampling, arXiv preprint arXiv:2209.07084
  (2022).

\bibitem{cao2022otkge}
Z.~Cao, Q.~Xu, Z.~Yang, Y.~He, X.~Cao, Q.~Huang, Otkge: Multi-modal knowledge
  graph embeddings via optimal transport, Advances in Neural Information
  Processing Systems 35 (2022) 39090--39102.

\bibitem{wang2019kgat}
X.~Wang, X.~He, Y.~Cao, M.~Liu, T.-S. Chua, Kgat: Knowledge graph attention
  network for recommendation, in: Proceedings of the 25th ACM SIGKDD
  international conference on knowledge discovery \& data mining, 2019, pp.
  950--958.

\bibitem{li2023imf}
X.~Li, X.~Zhao, J.~Xu, Y.~Zhang, C.~Xing, Imf: interactive multimodal fusion
  model for link prediction, in: Proceedings of the ACM Web Conference 2023,
  2023, pp. 2572--2580.

\bibitem{lee2023vista}
J.~Lee, C.~Chung, H.~Lee, S.~Jo, J.~Whang, Vista: Visual-textual knowledge
  graph representation learning, in: Findings of the Association for
  Computational Linguistics: EMNLP 2023, 2023, pp. 7314--7328.

\bibitem{zhang2024native}
Y.~Zhang, Z.~Chen, L.~Guo, Y.~Xu, B.~Hu, Z.~Liu, W.~Zhang, H.~Chen, Native:
  Multi-modal knowledge graph completion in the wild, in: Proceedings of the
  47th International ACM SIGIR Conference on Research and Development in
  Information Retrieval, 2024, pp. 91--101.

\bibitem{DBLP:conf/ijcai/XieLLS17}
R.~Xie, Z.~Liu, H.~Luan, M.~Sun,
  \href{https://doi.org/10.24963/ijcai.2017/438}{Image-embodied knowledge
  representation learning}, in: C.~Sierra (Ed.), Proceedings of the
  Twenty-Sixth International Joint Conference on Artificial Intelligence,
  {IJCAI} 2017, Melbourne, Australia, August 19-25, 2017, ijcai.org, 2017, pp.
  3140--3146.
\newblock \href {https://doi.org/10.24963/IJCAI.2017/438}
  {\path{doi:10.24963/IJCAI.2017/438}}.
\newline\urlprefix\url{https://doi.org/10.24963/ijcai.2017/438}

\bibitem{DBLP:conf/mm/ZhangQFX19}
Y.~Zhang, S.~Qian, Q.~Fang, C.~Xu,
  \href{https://doi.org/10.1145/3343031.3351033}{Multi-modal knowledge-aware
  hierarchical attention network for explainable medical question answering},
  in: L.~Amsaleg, B.~Huet, M.~A. Larson, G.~Gravier, H.~Hung, C.~Ngo, W.~T. Ooi
  (Eds.), Proceedings of the 27th {ACM} International Conference on Multimedia,
  {MM} 2019, Nice, France, October 21-25, 2019, {ACM}, 2019, pp. 1089--1097.
\newblock \href {https://doi.org/10.1145/3343031.3351033}
  {\path{doi:10.1145/3343031.3351033}}.
\newline\urlprefix\url{https://doi.org/10.1145/3343031.3351033}

\bibitem{limodality}
S.~Li, C.~Du, Y.~Huang, L.~Huang, H.~Zhao, Modality complementariness: Towards
  understanding multi-modal robustness.

\bibitem{liang2024survey}
K.~Liang, L.~Meng, M.~Liu, Y.~Liu, W.~Tu, S.~Wang, S.~Zhou, X.~Liu, F.~Sun,
  K.~He, A survey of knowledge graph reasoning on graph types: Static, dynamic,
  and multi-modal, IEEE Transactions on Pattern Analysis and Machine
  Intelligence (2024).

\bibitem{wang2017knowledge}
Q.~Wang, Z.~Mao, B.~Wang, L.~Guo, Knowledge graph embedding: A survey of
  approaches and applications, IEEE transactions on knowledge and data
  engineering 29~(12) (2017) 2724--2743.

\bibitem{yang2014embedding}
B.~Yang, W.-t. Yih, X.~He, J.~Gao, L.~Deng, Embedding entities and relations
  for learning and inference in knowledge bases, arXiv preprint arXiv:1412.6575
  (2014).

\bibitem{balavzevic2019tucker}
I.~Bala{\v{z}}evi{\'c}, C.~Allen, T.~M. Hospedales, Tucker: Tensor
  factorization for knowledge graph completion, arXiv preprint arXiv:1901.09590
  (2019).

\bibitem{wang2021visual}
M.~Wang, S.~Wang, H.~Yang, Z.~Zhang, X.~Chen, G.~Qi, Is visual context really
  helpful for knowledge graph? a representation learning perspective, in:
  Proceedings of the 29th ACM International Conference on Multimedia, 2021, pp.
  2735--2743.

\bibitem{xie2016image}
R.~Xie, Z.~Liu, H.~Luan, M.~Sun, Image-embodied knowledge representation
  learning, arXiv preprint arXiv:1609.07028 (2016).

\bibitem{zhang2024multimodal}
Q.~Zhang, Y.~Wei, Z.~Han, H.~Fu, X.~Peng, C.~Deng, Q.~Hu, C.~Xu, J.~Wen, D.~Hu,
  et~al., Multimodal fusion on low-quality data: A comprehensive survey, arXiv
  preprint arXiv:2404.18947 (2024).

\bibitem{wanyan2023active}
Y.~Wanyan, X.~Yang, C.~Chen, C.~Xu, Active exploration of multimodal
  complementarity for few-shot action recognition, in: Proceedings of the
  IEEE/CVF Conference on Computer Vision and Pattern Recognition, 2023, pp.
  6492--6502.

\bibitem{he2022new}
Y.~He, F.~Xiao, A new base function in basic probability assignment for
  conflict management, Applied Intelligence 52~(4) (2022) 4473--4487.

\bibitem{he2021conflicting}
Y.~He, F.~Xiao, Conflicting management of evidence combination from the point
  of improvement of basic probability assignment, International Journal of
  Intelligent Systems 36~(5) (2021) 1914--1942.

\bibitem{he2022mmget}
Y.~He, Y.~Deng, Mmget: a markov model for generalized evidence theory,
  Computational and Applied Mathematics 41 (2022) 1--41.

\bibitem{he2023tdqmf}
Y.~He, Y.~Deng, Tdqmf: Two-dimensional quantum mass function, Information
  Sciences 621 (2023) 749--765.

\bibitem{he2023ordinal}
Y.~He, Y.~Deng, Ordinal belief entropy, Soft Computing 27~(11) (2023)
  6973--6981.

\bibitem{he2022ordinal}
Y.~He, Y.~Deng, Ordinal fuzzy entropy, Iranian Journal of Fuzzy Systems 19~(3)
  (2022) 171--186.

\bibitem{he2024epl}
Y.~He, Epl: Evidential prototype learning for semi-supervised medical image
  segmentation, arXiv preprint arXiv:2404.06181 (2024).

\bibitem{he2024mutual}
Y.~He, Y.~Bi, L.~Li, C.-M. Pun, W.~Jiao, Z.~Jin, Mutual evidential deep
  learning for semi-supervised medical image segmentation, in: 2024 IEEE
  International Conference on Bioinformatics and Biomedicine (BIBM), IEEE,
  2024, pp. 2010--2017.

\bibitem{he2024uncertainty}
Y.~He, L.~Li, Uncertainty-aware evidential fusion-based learning for
  semi-supervised medical image segmentation, arXiv preprint arXiv:2404.06177
  (2024).

\bibitem{he2025co}
Y.~He, L.~Li, T.~Zhan, C.-M. Pun, W.~Jiao, Z.~Jin, Co-evidential fusion with
  information volume for semi-supervised medical image segmentation, Pattern
  Recognition 166 (2025) 111639.

\bibitem{li2024efficient}
L.~Li, Y.~He, C.-M. Pun, Efficient prototype consistency learning in
  semi-supervised medical image segmentation via joint uncertainty and data
  augmentation, in: 2024 IEEE International Conference on Bioinformatics and
  Biomedicine (BIBM), IEEE, 2024, pp. 2114--2121.

\bibitem{he2021matrix}
Y.~He, L.~Li, T.~Zhan, A matrix-based distance of pythagorean fuzzy set and its
  application in medical diagnosis, arXiv preprint arXiv:2102.01538 (2021).

\bibitem{li2025adaptive}
L.~Li, Y.~He, C.-M. Pun, An adaptive framework for multi-view clustering
  leveraging conditional entropy optimization, in: ICASSP 2025-2025 IEEE
  International Conference on Acoustics, Speech and Signal Processing (ICASSP),
  2025.

\bibitem{huang2025unitrans}
C.-j. Huang, Y.~He, X.~Han, W.~Jiao, Z.~Jin, L.~Wang, Unitrans: A unified
  vertical federated knowledge transfer framework for enhancing cross-hospital
  collaboration, arXiv preprint arXiv:2501.11388 (2025).

\bibitem{bi2025multi}
Y.~Bi, E.~Che, Y.~Chen, Y.~He, J.~Qu, Multi-prototype-based embedding
  refinement for medical image segmentation, in: ICASSP 2025-2025 IEEE
  International Conference on Acoustics, Speech and Signal Processing (ICASSP),
  IEEE, 2025, pp. 1--5.

\bibitem{li2025multi}
L.~Li, Z.~Jin, X.~Zhang, H.~Duan, J.~Wang, Z.~Tao, H.~Zhao, X.~Zhu, Multi-view
  riemannian manifolds fusion enhancement for knowledge graph completion, IEEE
  Transactions on Knowledge and Data Engineering (2025).

\bibitem{li2022nndf}
L.~Li, Y.~He, L.~Li, Nndf: A new neural detection network for aspect-category
  sentiment analysis, in: International Conference on Knowledge Science,
  Engineering and Management, Springer, 2022, pp. 339--355.

\bibitem{he2024generalized}
Y.~He, L.~Li, T.~Zhan, W.~Jiao, C.-M. Pun, Generalized uncertainty-based
  evidential fusion with hybrid multi-head attention for weak-supervised
  temporal action localization, in: ICASSP 2024-2024 IEEE International
  Conference on Acoustics, Speech and Signal Processing (ICASSP), IEEE, 2024,
  pp. 3855--3859.

\bibitem{xu2023spatio}
T.~Xu, K.~Yan, Y.~He, S.~Gao, K.~Yang, J.~Wang, J.~Liu, Z.~Liu, Spatio-temporal
  variability analysis of vegetation dynamics in china from 2000 to 2022 based
  on leaf area index: A multi-temporal image classification perspective, Remote
  Sensing 15~(12) (2023) 2975.

\bibitem{li2025rethinking}
L.~Li, Z.~Jin, Y.~He, D.~Jin, H.~Duan, Z.~Tao, X.~Zhang, J.~Li, Rethinking
  regularization methods for knowledge graph completion, arXiv preprint
  arXiv:2505.23442 (2025).

\bibitem{li2025towards}
L.~Li, Z.~Jin, Y.~Zhang, D.~Jin, C.~Dou, Y.~He, X.~Zhang, H.~Zhao, Towards
  structure-aware model for multi-modal knowledge graph completion, arXiv
  preprint arXiv:2505.21973 (2025).

\bibitem{chen2025revisit}
X.~Chen, Z.~Tao, K.~Zhang, C.~Zhou, W.~Gu, Y.~He, M.~Zhang, X.~Cai, H.~Zhao,
  Z.~Jin, Revisit self-debugging with self-generated tests for code generation,
  arXiv preprint arXiv:2501.12793 (2025).

\bibitem{simonyan2014very}
K.~Simonyan, A.~Zisserman, Very deep convolutional networks for large-scale
  image recognition, arXiv preprint arXiv:1409.1556 (2014).

\bibitem{zhang2024mixture}
Y.~Zhang, Z.~Chen, L.~Guo, Y.~Xu, B.~Hu, Z.~Liu, W.~Zhang, H.~Chen, Mixture of
  modality knowledge experts for robust multi-modal knowledge graph completion,
  arXiv preprint arXiv:2405.16869 (2024).

\bibitem{diederik2014adam}
P.~K. Diederik, Adam: A method for stochastic optimization, (No Title) (2014).

\bibitem{liu2019mmkg}
Y.~Liu, H.~Li, A.~Garcia-Duran, M.~Niepert, D.~Onoro-Rubio, D.~S. Rosenblum,
  Mmkg: multi-modal knowledge graphs, in: The Semantic Web: 16th International
  Conference, ESWC 2019, Portoro{\v{z}}, Slovenia, June 2--6, 2019, Proceedings
  16, Springer, 2019, pp. 459--474.

\bibitem{wang2023tiva}
X.~Wang, B.~Meng, H.~Chen, Y.~Meng, K.~Lv, W.~Zhu, Tiva-kg: A multimodal
  knowledge graph with text, image, video and audio, in: Proceedings of the
  31st ACM International Conference on Multimedia, 2023, pp. 2391--2399.

\bibitem{pan2022kuaipedia}
H.~Pan, Z.~Zhai, Y.~Zhang, R.~Fu, M.~Liu, Y.~Song, Z.~Wang, B.~Qin, Kuaipedia:
  a large-scale multi-modal short-video encyclopedia, arXiv preprint
  arXiv:2211.00732 (2022).

\bibitem{mousselly2018multimodal}
H.~Mousselly-Sergieh, T.~Botschen, I.~Gurevych, S.~Roth, A multimodal
  translation-based approach for knowledge graph representation learning, in:
  Proceedings of the Seventh Joint Conference on Lexical and Computational
  Semantics, 2018, pp. 225--234.

\bibitem{wang2019multimodal}
Z.~Wang, L.~Li, Q.~Li, D.~Zeng, Multimodal data enhanced representation
  learning for knowledge graphs, in: 2019 International Joint Conference on
  Neural Networks (IJCNN), IEEE, 2019, pp. 1--8.

\bibitem{lu2022mmkrl}
X.~Lu, L.~Wang, Z.~Jiang, S.~He, S.~Liu, Mmkrl: A robust embedding approach for
  multi-modal knowledge graph representation learning, Applied Intelligence
  (2022) 1--18.

\end{thebibliography}
\end{document}